\pdfoutput=1

\documentclass[11pt]{article}

\usepackage{ACL2023}

\usepackage{times}
\usepackage{latexsym}

\usepackage[T1]{fontenc}

\usepackage[utf8]{inputenc}

\usepackage{microtype}

\usepackage{inconsolata}
\usepackage{graphicx}
\usepackage{textcomp}
\usepackage{lipsum}
\usepackage{multirow}
\usepackage{pbox}
\usepackage{subcaption}
\usepackage{longtable}
\usepackage{array}

\title{PersonalityChat: Conversation Distillation \\ for Personalized Dialog Modeling with Facts and Traits}

\author{Ehsan Lotfi, Maxime De Bruyn, Jeska Buhmann, Walter Daelemans \\
        CLiPS Research Center \\ University of Antwerp, Belgium \\
        \texttt{firstname.lastname@uantwerpen.be}}

\begin{document}
\maketitle
\begin{abstract}
The new wave of Large Language Models (LLM) has offered an efficient tool to curate sizeable conversational datasets. So far studies have mainly focused on task-oriented or generic open-domain dialogs, and have not fully explored the ability of LLMs in following complicated prompts. In this work, we focus on personalization, and employ LLMs to curate a dataset which is difficult and costly to crowd-source: PersonalityChat is a synthetic conversational dataset based upon the popular PersonaChat dataset, but conditioned on both personas and (Big-5) personality traits. Evaluating models fine-tuned on this dataset, we show that the personality trait labels can be used for trait-based personalization of generative dialogue models. We also perform a head-to-head comparison between PersonalityChat and PersonaChat, and show that training on the distilled dataset results in more fluent and coherent dialog agents in the small-model regime.      

\end{abstract}

\section{Introduction}
Training deep learning models and architectures to act as decent dialog agents is a data-hungry task. Historically, big conversational datasets have been either curated from already existing dialog or sudo-dialog resources (e.g. Reddit \cite{dziri-etal-2019-augmenting}, Ubuntu \cite{lowe-etal-2015-ubuntu}, OpenSubtitles \cite{lison-tiedemann-2016-opensubtitles2016}), or via crowd-sourcing, which is usually the case if specific features are desired (e.g. persona-grounded \cite{zhang-etal-2018-personalizing}, empathetic \cite{rashkin-etal-2019-towards}). But these methods are noisy, costly or limited in the scope of feasible grounding features.

\begin{table}[t]
    \centering
    \begin{tabular}{|l|}
\hline
     \textbf{Persona:} \\ 
\hline
    I like to tan in tanning beds.\\
    I love spending money. \\
    I am high maintenance. \\
    I diet a lot. \\
    I love listening to Britney Spears. \\
\hline
\hline
    \textbf{Personality Traits (ChatGPT):} \hspace{35pt} \\
\hline
    \textbf{Openness}: No clues.\\
    \textbf{Conscientiousness}: Probably low, given the \\ 
     statement I love spending money and that  \\ 
     they admit to being high maintenance.\\ 
    \textbf{Extraversion}: Probably high, as they enjoy \\  
    tanning and listening to music, both of which \\ 
    are often done in social settings.\\ 
    \textbf{Agreeableness}: No clues.\\ 
    \textbf{Neuroticism}: Probably high, as they state\\   
    that they diet a lot and may have insecurities \\ 
    about their appearance. \\

\hline
    \end{tabular}
    
    \caption{Example of (Big-5) personality traits and explanations generated by ChatGPT when prompted with persona. We then sample the speculated trait labels, and use them along with personas to prompt ChatGPT for personalized dialogs.} 
    \label{tab:traits-expl}
\end{table} 

Recently, Large Language Models (LLMs) like GPT-3 \cite{NEURIPS2020_1457c0d6}, ChatGPT, GPT-4 and OPT \cite{zhang2022opt} provided a cheaper and more flexible alternative, which has been rapidly adopted to augment existing datasets \cite{chen2022weakly, mehri-etal-2022-lad, sahu-etal-2022-data} or generating new ones \cite{rosenbaum-etal-2022-linguist, 2022arXiv221210465K, Chen2023}. These studies have been mainly focused on task-oriented dialog, or scaling; i.e. exploiting the compelling efficiency and skill of the latest models to generate large-scale quality datasets that can help to mitigate some of the shortcoming of current dialog agents (e.g. common sense knowledge \cite{2022arXiv221210465K}). However the potential of these models in assuming sophisticated conversational settings has not been fully explored.      

In this study, we exploit this flexibility to distill a corpus which is difficult and expensive to curate from human interactions: a conversational dataset conditioned on both personas and the Big-5 personality traits. To prompt the LLM, we employ persona statements from the popular PersonaChat dataset \cite{zhang-etal-2018-personalizing} and enrich them with relevant labels from the well-known Big-5 personality traits \cite{https://doi.org/10.1111/j.1467-6494.1992.tb00970.x} which we also obtain using LLMs. Besides generating the first sizeable dialog corpus curated explicitly with personality traits, this parallel approach allows us to perform a fair comparison between the crowd-sourced PersonaChat and its distilled counterpart.

Our contributions are as follows: 1) We release \textbf{PersonalityChat}; a distilled personalized dialog dataset grounded on persona and personality traits. 2) We show that personality trait labels can be used to affect and modify the `attitude' of a dialog agent. 3) We demonstrate that in the small-model regime, training on PersonalityChat results in better performing dialog agents. 4) We also release \textbf{PersonaTraits}; a dataset of personality trait speculations, generated by ChatGPT for various personas\footnote{Datasets will be available at \url{https://github.com/ELotfi/PersonalityChat}.}.

\section{Related Work}
\subsection{Dialog Distillation from LLMs}
Early attempts to generate synthetic dialogues often employed schemes, agendas and templates to simulate data, mostly limited to task-oriented dialog \cite{schatzmann-etal-2007-agenda, shah-etal-2018-bootstrapping, campagna-etal-2020-zero}. Recently, the in-context zero-shot and few-shot learning capabilities of large language models, has made it possible to generate fluent customizable conversations either from scratch, or by providing a few examples. \citet{2022arXiv221210465K} prompted InstructGPT \cite{ouyang2022training} with simple narratives based on common sense statements to curate a large social dataset (SODA). \citet{Chen2023} used in-context learning, prompting the OPT \cite{zhang2022opt} model with a combination of topic, facts and dialog examples to generate another social dialog dataset extended to multi-party conversation (PLACES). \citet{lee-etal-2022-personachatgen} proposed a pipeline to augment and update PersonaChat with new personas using GPT-3.
More recently, MathDial \cite{macina2023mathdial} used a semi-synthetic framework to generate dialogs grounded in math reasoning problems by pairing real teachers with LLM-simulated students.

\subsection{Personalized Dialog Generation}
Before neural methods became mainstream, studies like \citet{mairesse-walker-2007-personage,Mairesse-2008,Mairesse2008CanCA} and \citet{gill-etal-2012-perceptions} used parameterised symbolic methods to incorporate the big-5 personality traits in dialog generation based on lexical features. In one of the first attempts for neural personalised dialog generation, \citet{li-etal-2016-persona} used jointly learned Speaker embeddings to inject personal information into the response. The release of PersonaChat \cite{zhang-etal-2018-personalizing} as the first large scale dataset grounded in personal facts, canonised the factual approach to dialog personalization. Curating a sizeable human-written dialog dataset grounded in personality traits is difficult and expensive. Recently \citet{saha-etal-2022-stylistic} exploited two trait-annotated non-conversational corpora (PANDORA \cite{gjurkovic-etal-2021-pandora} and Essays \cite{Pennebaker1999LinguisticSL}) to train a classifier which was used to annotate two existing knowledge-based datasets with trait labels. These annotations then were incorporated to train a personalised model.

\subsection{The Big-5 Personality Traits}
Rooted in the Lexical Hypothesis \cite{galton1884measurement} and developed over the last century, the Big Five personality traits \cite{Norman1963TowardAA, Goldberg1990AnA}, consist of Openness (to experience), Conscientiousness, Extraversion, Agreeableness and Neuroticism. These traits were repeatedly obtained empirically by extracting the main components of a factor analysis over adjective descriptors, and therefore considered as the dimensions explaining the most variance of behaviour among people \cite{Mairesse2008CanCA}. Although the Big Five model has limitations \cite{Paunonen2000WhatIB}, it has strong links to biological bases \cite{article-revelle} and is shown to be a stable personality evaluation metric \cite{COBBCLARK201211}. Table \ref{tab:trait-adje} in the Appendix (from \cite{https://doi.org/10.1111/j.1467-6494.1992.tb00970.x}) lists some of the defining adjective and attributes for the five dimensions.

\begin{figure*}[t]
    \centering
    \includegraphics[width=\textwidth]{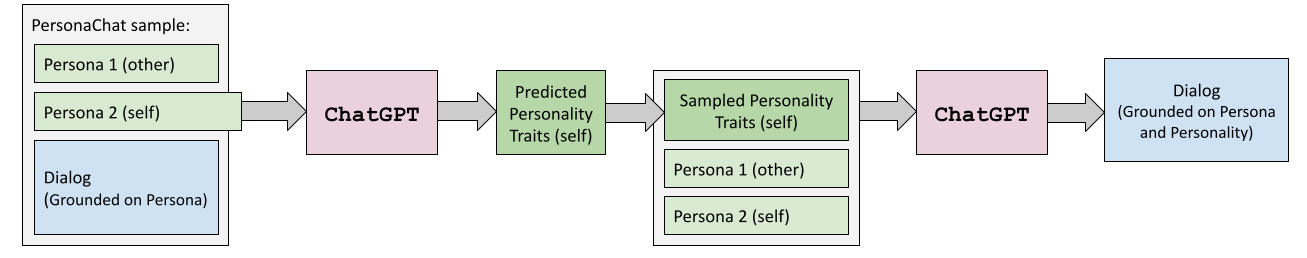}
    \caption{Overall pipeline to curate PersonalityChat. First (left) ChatGPT is used to predict personality traits for PersonaChat personas. Then (right) sampled traits (based on the predictions) are used along with the personas to prompt ChatGPT for dialogs that are grounded on both persona facts and personality traits.}
    \label{fig:pipe}
\end{figure*}

\section{PersonalityChat}
Generating a diverse set of conversations from LLMs is usually done using a varied collection of \textit{scenarios} or settings which provide the model with background information. For example \citet{2022arXiv221210465K} employed short narratives based on commonsense knowledge, and \citet{Chen2023} used a list of topics and tasks from the FITS dataset \cite{xu2022learning}. We base our work on PersonaChat \cite{zhang-etal-2018-personalizing}, the most popular dataset for dialog personalization which was curated by pairing online crowd-source workers and asking them to converse with each other using their assigned \textit{personas}; a small set of simple personal statements about occupation, interests, hobbies etc. (e.g. Table \ref{tab:traits-expl}-top). Using PersonaChat allows us to 1) exploit persona statements to acquire (estimated) personality traits, as well as dialog scenarios, 2) perform a head-to-head comparison between crowd-sourcing and distillation, as means of dialog dataset curation.           

The overall 2-step curation pipeline is shown in Figure \ref{fig:pipe}: first ChatGPT is employed to predict personality traits for personas from the PersonaChat dataset. Then traits are sampled from these labels and used along with the personas to prompt ChatGPT to generate dialogs grounded on both features. Next we describe these two steps in detail.

\subsection{Personality Trait Prediction}
\label{sec:trait_pred}
The PersonaChat dataset \cite{zhang-etal-2018-personalizing} comprises of 10,907 dialogs (8939/1000/968 for train/validation/test) between crowd-source workers. Dialog samples come with a pair of \textit{Persona}s labeled as `self' and `other', each containing 4-6 personal facts similar to what is shown in Table \ref{tab:traits-expl}-Top. Not all personas are unique and for example the `self' personas are permuted from 4333/409/414 unique sets.

\begin{figure}[ht]
    \centering
    \includegraphics[width=.5\textwidth]{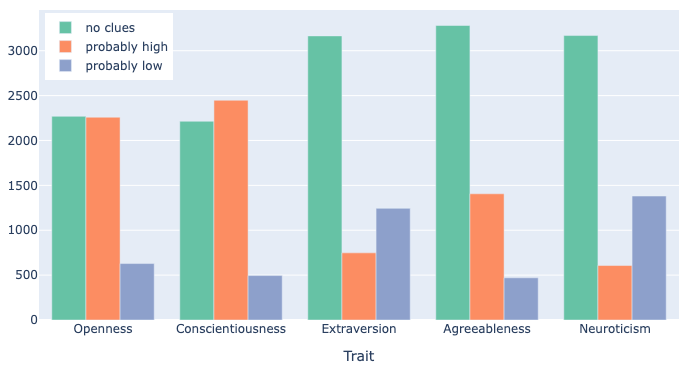}
    \caption{Distribution of the estimated trait labels for PersonaChat personas}
    \label{fig:trait-distr}
\end{figure}

Personas provide enough background information to prompt LLMs for social dialog and in theory we can just enrich this information with randomly sampled personality traits to generate our prompts. But since persona facts and personality traits are not independent, in order to better align our (trait) sampling with personas, we first use ChatGPT to acquire trait label estimations for each persona, as one of `probably low', `probably high' or `no clues'. (The exact prompt, and examples of the predicted labels and supporting explanations can be found in Appendix \ref{sec:appendix_aa}) and Table \ref{tab:trait-exmp} respectively). Figure \ref{fig:trait-distr} shows the trait-label distribution, indicating that in 4 out of 5 traits, `no clues' is the majority label. On average personas have 2.3 \textit{indicative} labels (i.e. 'probably low' or 'probably high')\footnote{We release the acquired dataset as \textbf{PersonaTraits}, including 5156 samples of persona, 3-way labels and explanations.}.   

\begin{table}[h]
    \centering
    \begin{tabular}{|l|>{\centering\arraybackslash}p{1.5cm}|>{\centering\arraybackslash}p{1.5cm}|>{\centering\arraybackslash}p{1.6cm}|}
    \hline
     \small Method & \small Ave. \# indicative labels & \small Ave. Hamming distance  & \small Ave. polar Hamming distance \\
    \hline
    \small Human & $2.35 \pm .14$ & - & - \\
    \small ChatGPT & $2.27$ & $1.88 \pm .06$ & $.13 \pm .01$ \\
    \small random & $3.45$ & $3.42 \pm .1$ & $.85 \pm .04$ \\
    \hline       

    \end{tabular}
    \caption{Comparing human annotations for persona-based traits with ChatGPT and random labels on 100 samples. Distances are measured from the human annotations. Polar Hamming distance only considers polar or extreme disagreement; i.e. high vs. low.  } 
    \label{tab:ptrait_annot}
\end{table}  

Predicting trait labels from a few facts is inherently noisy; in most cases persona statements do not \textit{strongly} imply or dictate a specific trait, which means that the actual label is a broad distribution. Since we only use these labels for sampling to prompt the LLM (as an alternative to random labels), what is relevant to us is to make sure that: 1) the labels are in fact closer to human annotations than random sampling, 2) they have relatively low polar disagreements (`high' vs. `low' for a trait) with the human annotation. To do so, we ask two expert annotators to label a set of 100 personas with the same categories (`probably low', `probably high' or `no clues'). Table \ref{tab:ptrait_annot} shows how these annotations compare with ChatGPT and random labels. The difference is measured using the Hamming distance (for five traits), and the polar distance only considers extreme disagreement. As can be seen, ChatGPT labels are clearly closer to human annotations, and while the exact agreement is not very high, the polar Hamming distance is quite low (.13 on average). In the next step, we use personas and their acquired trait labels to prompt ChatGPT for social conversations.

\subsection{Personalized Dialog Generation}
\subsubsection{Sampling Trait Labels}
Although we can use labels as they are, we choose to sample based on them to control the complexity, as some personas have 4 or 5 indicative personality labels, which can impose \textit{too many} constraints for a relatively short dialog. In addition, since we are using the same model (ChatGPT) for both steps, we speculate that introducing some randomness would smooth potential lexical links and biases between personas and their estimated traits, and encourage the model to explore more sophisticated ways to reflect the traits in utterances, rather than relying solely on the relevant personal facts. Therefore, for each persona with $k$ indicative trait-labels (i.e. `high' or `low'), we sample $n$ ($0<n<4$) indicative labels, following these rules: 
\begin{itemize}
    \item If $n <= k$, we randomly pick $n$ labels.
    \item Otherwise, we compensate the difference by randomly switching $n - k$ non-indicative labels (i.e. `no clues') to an indicative one.   
\end{itemize}

\subsubsection{Prompting}
Nest step is prompting ChatGPT with personas and personality traits for grounded dialogs. This can be done with `single agent' --where one prompt is made with information for both speakers and the model is asked to generate a conversation between them in one pass--, or `double agent' where each speaker gets their own API (prompted with corresponding information) and the conversation is generated one utterance at a time going back and forth between the agents with updated contexts. In theory the double agent approach sounds more faithful to the real dynamics of a conversation\footnote{As well as the crowd-sourcing method in PersonaChat} but according to our trials, it often triggers the `assistant' side of ChatGPT (resulting in long and `helpful' utterances!) which is hard to mitigate with prompts. Moreover, it is not easy to anticipate or control the length of conversation as it might `end' before the generation loop is over. The single agent approach on the other hand is simpler and cheaper (on average consuming almost 1/10 tokens) but it is prone to persona `leakage'; i.e. one speaker referring to the other speaker's information. 
Another challenge (in both generation settings) is avoiding \textit{explicit} grounding; i.e. cases where the speaker talks about their personality traits in a self-aware and superficial manner (e.g. ``I am a very conscientious person.''), instead of implicitly reflecting them in responses. We realize that both these issues, can be considerably mitigated by prompt engineering, but not totally eradicated.

After many trials, we opt for the single agent method, plus a lexical post processing step to remove (and regenerate) the most `leaky' and `explicit' cases. To control the complexity, we only provide personality traits for the second speaker, corresponding to the `self' persona (or Person2 in our terminology). We generate a dialog for each PersonaChat sample (i.e. persona pair), picking the temperature uniformly from the $(.25,1.25)$ interval and limiting the number of generated tokens to 450. The exact prompt can be found in Appendix \ref{sec:appendix_aa}.

\subsection{Dataset Statistics}
\label{sec:stat}
\subsubsection{Lexical Features}
\label{sec:lex}
Table \ref{tab:compr} compares various lexical properties of the resulted dataset (PersonalityChat) with those of PersonaChat. \textbf{MTLD} is the Measure of Textual Lexical Diversity \cite{McCarthy2010MTLDVA}, calculated as the mean length of sequential word strings in a text that maintain a given TTR value\footnote{We use the same TTR value (.72) as employed by the authors \cite{McCarthy2010MTLDVA}. For the rationale see the paper.}, and can be interpreted as an indication for local diversity. Based on the numbers, PersonalityChat has a smaller vocabulary\footnote{The fact that PersonaChat is not normalized and contains erroneous and non-standard spellings contributes to this difference.} but a higher MTLD score, which points to a discrepancy in how the inter- and intra-dialog diversity compare between the two datasets. 
To see how verbatim the persona facts have been employed in dialogs, we measure the average maximal common n-gram length (\textbf{MCN}) between persona facts and utterances or whole dialogs\footnote{For example MCN=3 for dialog means that on average, the longest n-gram shared between individual persona facts and the corresponding dialog has a length of 3 (tokens).}, which results in a higher value for PersonalityChat, specially over the whole conversation. 

The dialog example in Table \ref{tab:exmpl-chats} demonstrates some aspects of the lexical and stylistic differences: The language usage in the crowd-sourced example (PersonaChat) is more creative, specially in the way that persona facts are being incorporated. This however comes with a significantly higher variance, and while synthesized dialogs almost always follow a safe and streamlined style of fact-grounded personal conversation, the crowd-sourced samples vary between exemplary ones like Table \ref{tab:exmpl-chats}, and low quality cases where persona facts are just exchanged with little or no conversational context.

\subsubsection{Curation Cost}
Table \ref{tab:cost} shows the generation cost for both steps, i.e. personality trait prediction, and dialog generation, which overall amounts to 15\$ and 70hrs. This excludes trials and corrective re-generations which in our case add an extra 5 \%.

\begin{table}[t]
    \centering
    \begin{tabular}{|l|c|c|}
    \hline
      & \small PersonaChat & \small PersonalityChat \\
    \hline       
    Dialogs & 10,907 & 10,907 \\ 
    Turns \small(avg. dialog)  & 14.8 & 17.3\\
    Tokens \small(avg. turn) & 11.7 & 13.7 \\ 
    MTLD & 50.3 & 69.1 \\
    MCN \small(dialog) & 1.76 & 3.38 \\
    MCN \small(turn) & 1.23 & 2.0 \\
    Vocab size \small(train) &18,669 &11,104  \\
    
    \hline
    \end{tabular}
    \caption{Lexical comparison between PersonaChat \cite{zhang-etal-2018-personalizing} and our distilled PersonalityChat dataset. MTLD stands for Measure of Textual Lexical Diversity \cite{McCarthy2010MTLDVA} and MCN is the average maximal common n-gram between persona and dialog (or turn), as described in \ref{sec:stat}.} 
    \label{tab:compr}
\end{table}  

\begin{table}[h]
    \centering
    \begin{tabular}{|l|c|c||c|c||c|c|}
    \hline
      & \multicolumn{2}{c||}{Step 1} & \multicolumn{2}{c||}{Step 2} & \multicolumn{2}{c|}{Overall}\\
      & ps & all & ps & all & ps & all \\
    \hline       
    Time     & 6s & 8.6h & 20s & 61h & 23s & 70h \\  
    Cost & .05\textcent & 2.6\$ & .1\textcent & 12\$ & .13\textcent & 15\$ \\
    \hline
    \end{tabular}
    \caption{Curation cost for both steps in our method; i.e. personality trait prediction (step 1) and dialog generation (step 2). `ps' is the `per sample' value, which does not add up to the last column since each persona is used in almost two dialogs.} 
    \label{tab:cost}
\end{table}

\section{Training with PersonalityChat}
\label{sec:train}
In this section we perform experiments to answer the following questions:
\begin{itemize}
    \item \textbf{RQ1}: Can we use personality traits to control the conversational behavior of a model?
    \item \textbf{RQ2}: How do the models trained on PersonaChat and PersonalityChat compare in performance? (only using persona)
\end{itemize}

\subsection{RQ1: Controlling Personality Traits}
\label{sec:q2}
To answer the first question, we fine-tune a BART model\footnote{The large version accessible via HuggingFace's \texttt{transformers} as \texttt{facebook/bart-large}} \cite{lewis-etal-2020-bart} on our dataset, using both personas and personality traits in the context. Following the common practice\footnote{Using specific tokens and phrases to control language generation is inspired by similar works, specially CTRL \cite{Keskar2019CTRLAC}.}, we compose the inputs by concatenating different segments as \texttt{<traits> \{trait labels\} <persona> \{persona facts\} \{history\}}, where \texttt{history} includes previous turns marked by \texttt{<user>} and \texttt{<agent>}, and \texttt{<traits>}, \texttt{<persona>}, \texttt{<user>} and \texttt{<agent>} are special tokens added to the vocabulary. We do the fine-tuning using HuggingFace's \texttt{transformers} library \cite{wolf-etal-2020-transformers} for 8 epochs, while the validation loss declines (More details in Appendix \ref{sec:appendix_b}).

To see if the trained models are capable of generating trait-adapted responses, we perform two sets of evaluations:

\begin{figure*}[ht]
    \centering
    \includegraphics[width=\textwidth]{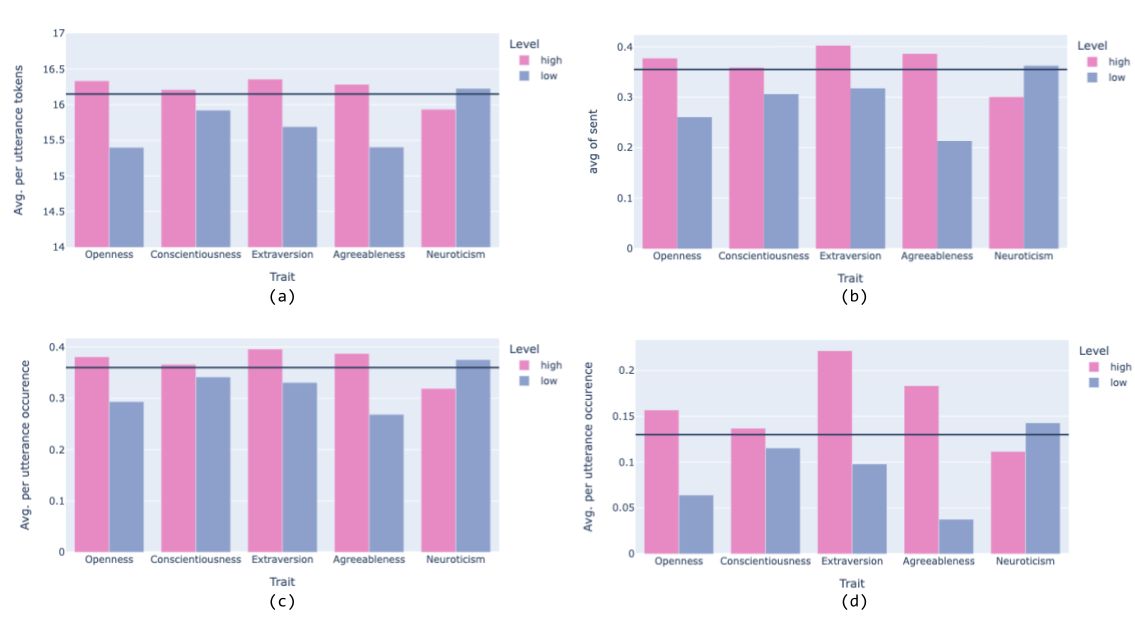}
    \caption{Comparing single-trait utterances based on simple lexical/semantic features: a) number of tokens,  b) sentiment score,  c) number of question marks, d) number of exclamation marks. The horizontal black line indicates the value in utterances generated without specifying any traits. Values are average over utterances.}
    \label{fig:trait-lex-comp}
\end{figure*}

\begin{table*}[h]
    \centering
    \begin{tabular}{|l|c|c|c||c|c|c||c|c|c||c|c|c||c|c|c|}
    \hline
    & \multicolumn{3}{|c||}{\small Openness} & \multicolumn{3}{c||}{\small Conscientiousness} & \multicolumn{3}{c||}{\small Extraversion} & \multicolumn{3}{c||}{\small Agreeableness} & \multicolumn{3}{c|}{\small Neuroticism}\\
   \small Model & \small high & \small tie & \small low & \small high & \small tie & \small low & \small high & \small tie & \small low & \small high & \small tie & \small low & \small high & \small tie & \small low \\
    \hline       
    \small Ave. vote & \small66\tiny\% &\small 18\tiny\% & \small 16\tiny\% &\small 36\tiny\% &\small 40\tiny\% &\small 24\tiny\%     &\small 64\tiny\% &\small 20\tiny\% &\small16\tiny\% &\small 72\tiny\% &\small 14\tiny\% &\small 14\tiny\% &\small42\tiny\%        &\small46\tiny\% &\small14\tiny\%\\  
    \hline
    \small Cohen's $\kappa$ & \multicolumn{3}{|c||}{\small0.56} & \multicolumn{3}{c||}{\small0.29} & \multicolumn{3}{c||}{\small0.46} & \multicolumn{3}{c||}{\small0.49} & \multicolumn{3}{c|}{\small0.25}\\
    \hline
    \end{tabular}
    \caption{ Pairwise per-dialog comparison by human annotators between dialogs generated using opposing trait labels (e.g. high and low openness). The second row shows Cohen's kappa between annotators.} 
    \label{tab:trait_comp}
\end{table*}

\subsubsection{Turn-wise Automatic Evaluation}
For this part, we use the whole validation set (\texttildelow 8k samples) and generate responses using single trait labels, or no labels (i.e. $5*2+1=11$ responses for each sample, the last of which we consider as `neutral'). Then we compare a number of indicative lexical/semantic features between the inferred utterances, including average sentiment score, average number of tokens, question marks and exclamation marks per utterance. For sentiment we employ a BERT-base model finetuned on the 5-way review classification and use the class probabilities to calculate an aggregated score\footnote{More specifically we calculate $\Sigma_{i=0}^{4} s[i]*p_{i}$ where $s=[-1, -.5, 0, .5, 1]$.}.  The result (Figure \ref{fig:trait-lex-comp}) shows a consistent pattern in which utterances generated with higher openness, extraversion, agreeableness and conscientiousness, or lower neuroticism are more expressive (a and d), positive (b) and engaging (c), with the difference being relatively higher for openness, extraversion and agreeableness. It also shows that overall, the `low' label is more effective in steering the response away from the neutral behavior (i.e. the no-trait baseline marked with a black line in each graph).

To have a more semantic view, we identify the frequently `preferred' and `avoided' n-grams for each trait-label by comparing n-gram frequencies in the trait and neutral utterances. The result ( Table \ref{tab:trait_ngram} in the Appendix) shows that the statistically indicative n-grams for each trait-label combination\footnote{This is measured by $abs(x-y)/y$ conditioned on $abs(x-y)>20$ where $x$ and $y$ are the occurrence frequency in the trait and neutral corpora, respectively.} generally agree with common semantic expectations for that trait-label (cf. Table \ref{tab:trait-adje}), although the diversity and richness varies quite noticeably between different traits and labels. In particular, the `low neuroticism' corpora does not offer any particularly important n-grams.

\begin{table*}[ht]
    \centering
    \begin{tabular}{|l||c|c|c|c|c||c|c|c|c|c||c|c|c|c|c|}
    \hline
     \hspace{6pt} \textbackslash \small Eval & \multicolumn{5}{c||}{\small PChat} & \multicolumn{5}{c||}{\small PTChat} & \multicolumn{5}{c|}{\small Both}\\
    \small Train \textbackslash  & \small PPL & \small F1 & \small TTR & \small MCN & \small H  & \small PPL & \small F1 & \small TTR & \small MCN & \small H & \small PPL & \small F1 & \small TTR & \small MCN & \small H \\
    \hline       
    \small PChat     & \small 9.88 &\small  .18 &\small  .71 &\small  4.75 &\small  1.3 &\small  21.5 & \small .21 & \small .76 & \small 3.58 & \small 1.2 & \small 15.7 & \small .19 & \small .73 & \small 4.20 & \small 1.3 \\  
    \small PTChat     & \small 56.3 & \small .17 & \small .92 & \small 4.20 & \small 2.0 & \small 3.52 & \small .30& \small .91 & \small 3.44 & \small 2.2 & \small 10.8 & \small .24 & \small .92 & \small 3.81 & \small 2.1 \\ 
    \small Both       & \small 9.69 & \small .19 & \small .80 & \small 4.15 & \small 1.8 & \small 3.40 & \small .31 & \small .92 & \small 3.14 & \small  2.3 & \small 5.21 & \small .25 & \small .86 & \small 3.65 & \small 2.1 \\ 
    \hline
    \end{tabular}
    \caption{Performance metrics for various train (rows)-evaluation (columns) settings, using \textbf{T5-small}. PPL and TTR are perplexity and type/token ratio. MCN is the average length of the maximal n-gram shared between response and persona facts, and H is overall score (from 1 to 3) assigned by human annotators.} 
    \label{tab:pc-vs-pt}

\vspace{1em}

    \centering
    \begin{tabular}{|l|c|c|c||c|c|c||c|c|c||c|c|c||c|c|c|}
    \hline
    & \multicolumn{3}{|c||}{\small Naturalness}  & \multicolumn{3}{c||}{\small Coherence }  & \multicolumn{3}{c||}{\small Consistency }  & \multicolumn{3}{c||}{\small Specificity }   & \multicolumn{3}{c|}{\small Overall } \\
   \small Domain & \small PT & \small tie & \small P & \small PT & \small tie & \small P & \small PT & \small tie & \small P & \small PT & \small tie & \small P & \small PT & \small tie & \small P \\
    \hline       
    \small PChat & \small70\tiny\% &\small 26\tiny\% & \small 4\tiny\% &\small 49\tiny\% &\small 40\tiny\% &\small 11\tiny\%     &\small 25\tiny\% &\small 64\tiny\% &\small11\tiny\% &\small 12\tiny\% &\small 67\tiny\% &\small 21\tiny\% &\small74\tiny\%        &\small15\tiny\% &\small11\tiny\%\\  
    \small PTChat & \small66\tiny\% &\small 34\tiny\% & \small 0\tiny\% &\small 64\tiny\% &\small 36\tiny\% &\small 0\tiny\%      &\small 29\tiny\% &\small 62\tiny\% &\small10\tiny\% &\small 13\tiny\% &\small 70\tiny\% &\small 17\tiny\% &\small81\tiny\%         &\small17\tiny\% &\small2\tiny\%\\
    \hline
    \small Ave. & \small68\tiny\% &\small 30\tiny\% & \small 2\tiny\% &\small 56\tiny\% &\small 38\tiny\% &\small 6\tiny\%        &\small 26\tiny\% &\small 64\tiny\% &\small10\tiny\% &\small 12\tiny\% &\small 69\tiny\% &\small 19\tiny\% &\small77\tiny\%     &\small16\tiny\% &\small7\tiny\%\\
    \hline
    \end{tabular}
    \caption{Head-to-head quality comparison by human annotators between the output of \textbf{T5-small} models fine-tuned on PersonaChat (P), and PersonalityChat (PT). Rows indicate the test data domain.} 
    \label{tab:t5small-comp-pt-vs-pc}

\vspace{2em}

    \centering
    \begin{tabular}{|l||c|c|c|c|c||c|c|c|c|c||c|c|c|c|c|}
    \hline
     \hspace{6pt} \textbackslash \small Eval & \multicolumn{5}{c||}{\small PChat} & \multicolumn{5}{c||}{\small PTChat} & \multicolumn{5}{c|}{\small Both}\\
    \small Train \textbackslash  & \small PPL & \small F1 & \small TTR & \small MCN & \small H  & \small PPL & \small F1 & \small TTR & \small MCN & \small H & \small PPL & \small F1 & \small TTR & \small MCN & \small H \\
    \hline       
    \small PChat     & \small 7.56 &\small  .21 &\small  .85 &\small  3.82 &\small  2.3 &\small  13.9 & \small .25 & \small .88 & \small 2.50 & \small 2.1 & \small 10.9 & \small .23 & \small .87 & \small 3.16 & \small 2.2 \\  
    \small PTChat     & \small 52.0 & \small .18 & \small .94 & \small 3.36 & \small 2.7 & \small 2.60 & \small .36 & \small .94 & \small 2.95 & \small 2.8 & \small 8.81 & \small .27 & \small .94 & \small 3.15 & \small 2.8 \\ 
    \small Both       & \small 7.52 & \small .21 & \small .88 & \small 3.40 & \small 2.4 & \small 2.55 & \small .38 & \small .94 & \small 2.79 & \small  2.7 & \small 3.96 & \small .30 & \small .91 & \small 3.10 & \small 2.6 \\ 
    \hline
    \end{tabular}
    \caption{Performance metrics for various train (rows)-evaluation (columns) settings, using \textbf{T5-base}.} 
    \label{tab:pc-vs-pt-base}

\vspace{1em}

    \centering
    \begin{tabular}{|l|c|c|c||c|c|c||c|c|c||c|c|c||c|c|c|}
    \hline
    & \multicolumn{3}{|c||}{\small Naturalness} & \multicolumn{3}{c||}{\small Coherence} & \multicolumn{3}{c||}{\small Consistency} & \multicolumn{3}{c||}{\small Specificity} & \multicolumn{3}{c|}{\small Overall}\\
   \small Domain & \small PT & \small tie & \small P & \small PT & \small tie & \small P & \small PT & \small tie & \small P & \small PT & \small tie & \small P & \small PT & \small tie & \small P \\
    \hline       
    \small PChat & \small23\tiny\% &\small 73\tiny\% & \small 4\tiny\% &\small 40\tiny\% &\small 49\tiny\% &\small 11\tiny\%     &\small 19\tiny\% &\small 79\tiny\% &\small2\tiny\% &\small 11\tiny\% &\small 75\tiny\% &\small 13\tiny\% &\small60\tiny\%        &\small15\tiny\% &\small25\tiny\%\\  
    \small PTChat & \small20\tiny\% &\small 80\tiny\% & \small 0\tiny\% &\small 64\tiny\% &\small 30\tiny\% &\small 6\tiny\%      &\small 21\tiny\% &\small 72\tiny\% &\small7\tiny\% &\small 19\tiny\% &\small 79\tiny\% &\small 2\tiny\% &\small79\tiny\%         &\small10\tiny\% &\small11\tiny\%\\
    \hline
    \small Ave. & \small21\tiny\% &\small 77\tiny\% & \small 2\tiny\% &\small 52\tiny\% &\small 40\tiny\% &\small 9\tiny\%        &\small 20\tiny\% &\small 76\tiny\% &\small4\tiny\% &\small 15\tiny\% &\small 77\tiny\% &\small 8\tiny\% &\small69\tiny\%     &\small13\tiny\% &\small18\tiny\%\\
    \hline
    \end{tabular}
    \caption{Head-to-head quality comparison by human annotators between the output of \textbf{T5-base} models fine-tuned on PersonaChat (P) and PersonalityChat (PT). Rows indicate the test data domain.} 
    \label{tab:t5base-comp}
\end{table*}

\subsubsection{Pairwise Per-dialog Human Evaluation}
\label{sec:rq2_eval}
To see if the observed statistical patterns lead to discernible semantic differences during interaction, we conduct a human evaluation. Following \citet{smith-etal-2022-human} who argue for the advantage of per-dialog comparison (vs. per-turn) in situations where differences between models only emerge after several conversation turns, we simulate 100 full conversations with random lengths between two models, one as user (trained on the `odd' utterances without traits), and one as agent (trained on the `even' utterances with traits), using persona pairs from the validation set. To mitigate the persona-trait bias, we filter the persona pool, keeping only personas with maximum 2 indicative trait labels (i.e. labels other than `no clues'. See \ref{sec:trait_pred}). For each sampled persona pair, we generate a pair of conversations for every trait, one with `high' and one with `low' label. This results in 500 dialog pairs, generated with single opposing trait labels.

To evaluate the outcome, we ask two annotators ---one psychology student familiar with the topic, and one in-house annotator provided with generic explanatory material on the Big-5 traits--- to indicate in which dialog of a pair, the agent shows a \textit{higher} value for that trait, with a possibility for a tie. Table \ref{tab:trait_comp} shows the average vote, along with annotator's agreement (Cohen’s kappa) for each trait, based on which we can conclude that the model is capable of noticeably reflecting Openness, Extraversion and Agreeableness in its output, while Conscientiousness and Neuroticism are not incorporated enough. This is consistent with the automatic results (Figure \ref{fig:trait-lex-comp}). Table \ref{tab:trait_dialog} (in the Appendix) shows 5 pairs from our evaluation set, cherry-picked to demonstrate how the model's outcome is affected by trait labels.


\subsection{RQ2: PersonaChat vs. PersonalityChat}
\label{sec:q1}
To investigate the second question, we fine-tune sequence-to-sequence pretrained transformers on these data (without the trait labels), and compare the resulting models using automatic metrics as well as human evaluation. Since we are interested in the low-parameter regime, we first select T5-small\footnote{More precisely, we use T5-v1.1-small which is an improved version.} which is one of the smallest pretrained encoder-decoders with around 77M parameters. For a fair comparison, we train on the odd utterances (first speaker), since their responses are not grounded in personality traits in our dataset. For brevity, we will refer to PersonaChat and PersonalityChat as \textbf{PChat} and \textbf{PTChat} respectively, in the rest of this section.

To evaluate and compare the models, we infer responses from them for the PChat and/or PTChat validation set, using a beam-search of size 5. For automatic metrics, we employ \textbf{PPL} (validation perplexity), \textbf{F1} (F1 measure between generated and gold response), \textbf{TTR} (type/token ratio) and \textbf{MCN} (described in \ref{sec:lex}). We also ask in-house human annotators to score a 200-sample subset of generated responses (100 for each domain) from 1 to 3, in terms of overall quality\footnote{3 being a response that is fluent, coherent, consistent and specific. More details in Appendix \ref{sec:appendix_c}.} (\textbf{H} column). 

Table \ref{tab:pc-vs-pt}, summarizes the results for 9 different settings, depending on the train (rows) and evaluation (columns) domains, which can be PChat, PTChat or both.
As can be seen, even with T5-small the in-domain perplexity of the PTChat model drops to the very low value of 3.52 (as opposed to 9.88 for PChat model). In terms of generation quality, training on PTChat (second row) generally results in better metrics, even on the out-of-domain PChat data. In particular, the PTChat model (second row) generates significantly less repetitive responses than the PChat model (higher TTR), and acquires a lower MCN score, which is interesting considering the higher MCN value in the PTChat dataset (Table \ref{tab:compr}). Table \ref{tab:pc-vs-pt} also shows that training on both datasets (last row) improves the PChat performance more than PTChat (PChat column: row 3 vs. row 1, compared to PTChat column: row 3 vs. row 2).

To have a more detailed result, we perform head-to-head comparison between the outputs of these models, from the same 200 samples: We ask our annotators to pick the `better' response in terms of Naturalness, Coherence, Consistency, Specificity and Overall quality, with a possibility for a tie\footnote{More details in Appendix \ref{sec:appendix_c}.}. Table \ref{tab:t5small-comp-pt-vs-pc} shows that the PT model (trained on PersonalityChat) beats the P model (trained on PersonaChat) on both domains in the overall preference, as well as specific features, except for Specificity. In particular, PT generates significantly more natural and coherent responses than P. Table \ref{tab:t5small-comp-pt-vs-pct} shows this advantage still holds in some categories over the PPT model (trained on both datasets), although with a narrower margin.

To see whether these results generalize to bigger models, we repeat the experiments using T5-base (\texttildelow 220M parameters). The results (Tables \ref{tab:pc-vs-pt-base}, \ref{tab:t5base-comp} and \ref{tab:t5base-comp-pt-vs-pct}) show that while PChat benefits more from the increase in model size, PTChat still holds its overall advantage.

\section{Conclusion}
In this work we exploited the highly customizable generation skills of ChatGPT to build a corpus which is difficult and expensive to curate from human interaction; i.e. a sizeable dialog dataset grounded in both personal facts and (Big-five) personality traits. Starting from the PersonaChat dataset, we enriched the persona statements with personality trait speculations and used them to prompt ChatGPT for personal dialog.
We explored the qualities of this dataset in two parts: 1) trait modification, which showed that the overall `attitude' of a dialog agent trained on our dataset can be modified using trait labels, specially for Openness, Extraversion and Agreeableness. 2) comparison with PersonaChat (using only persona facts) made possible by the parallel one-to-one curation scheme, which showed our distilled dataset can be used to train smaller dialog models with better performance, specially in terms of naturalness and coherence.  

We also faced and discussed some of the potential shortcomings of our approach. Most importantly, the `easy-training' advantage of a distilled dataset comes with a less diverse and more predictable language distribution which can make a dialog agent sound `boring' on the long run. Combining the natural and synthetic datasets can be a solution but (at least in our experiments) the distilled distribution seems to mostly overrule the natural one. In trait modification, we observed interesting results in the way trait labels can affect and steer model's conversational behavior but there is a lot of room for improvement (e.g. Table \ref{tab:trait_comp}). One idea for example might be making use of the explanatory PersonaTraits dataset, which we only released as a side product in this study.

\section*{Limitations}
The first limitation of our work regards various biases that are incorporated in the curated data as the result of our choices. Most importantly ChatGPT introduces its (regularised) preferences, specially in dialog generation. Employing a non-proprietary model would have given us more control over the generation process but this comes with significantly less flexibility, and a necessary post-processing step to check for outliers and harmful content. On a more subtle level, using PersonaChat personas to seed personality traits is not ideal since their generally expressive content and cold/neutral tone, bears implications regarding personality traits which affects the trait distribution.       

Our evaluation process, although standard and relatively comprehensive, has limitations which are mostly common in dialog assessment. The automatic metrics are calculated over responses generated from validation or test context which does not necessarily generalise to the real world interaction with an agent. More particular to our work, is the way we evaluated trait modification by using only one trait label at a time, which although makes calculations easier and more reliable, does not necessarily reflect the model's behavior when prompted with multiple trait labels. The potential persona-personality correlation is another influencing parameter in this step, which although we tried to mitigate by using more neutral persona sets, is not completely resolved.         

Finally and as mentioned in the previous section, the trait incorporation in generated dialogs is still basic and rather superficial, which only leaves more room for future studies and improvements.


\section*{Acknowledgements}
This research received funding from the Flemish Government under the “Onderzoeksprogramma Artificiële Intelligentie (AI) Vlaanderen” programme.

\bibliography{anthology,custom}

\begin{thebibliography}{40}
\expandafter\ifx\csname natexlab\endcsname\relax\def\natexlab#1{#1}\fi

\bibitem[{Brown et~al.(2020)Brown, Mann, Ryder, Subbiah, Kaplan, Dhariwal, Neelakantan, Shyam, Sastry, Askell, Agarwal, Herbert-Voss, Krueger, Henighan, Child, Ramesh, Ziegler, Wu, Winter, Hesse, Chen, Sigler, Litwin, Gray, Chess, Clark, Berner, McCandlish, Radford, Sutskever, and Amodei}]{NEURIPS2020_1457c0d6}
Tom Brown, Benjamin Mann, Nick Ryder, Melanie Subbiah, Jared~D Kaplan, Prafulla Dhariwal, Arvind Neelakantan, Pranav Shyam, Girish Sastry, Amanda Askell, Sandhini Agarwal, Ariel Herbert-Voss, Gretchen Krueger, Tom Henighan, Rewon Child, Aditya Ramesh, Daniel Ziegler, Jeffrey Wu, Clemens Winter, Chris Hesse, Mark Chen, Eric Sigler, Mateusz Litwin, Scott Gray, Benjamin Chess, Jack Clark, Christopher Berner, Sam McCandlish, Alec Radford, Ilya Sutskever, and Dario Amodei. 2020.
\newblock \href {https://proceedings.neurips.cc/paper_files/paper/2020/file/1457c0d6bfcb4967418bfb8ac142f64a-Paper.pdf} {Language models are few-shot learners}.
\newblock In \emph{Advances in Neural Information Processing Systems}, volume~33, pages 1877--1901. Curran Associates, Inc.

\bibitem[{Campagna et~al.(2020)Campagna, Foryciarz, Moradshahi, and Lam}]{campagna-etal-2020-zero}
Giovanni Campagna, Agata Foryciarz, Mehrad Moradshahi, and Monica Lam. 2020.
\newblock \href {https://doi.org/10.18653/v1/2020.acl-main.12} {Zero-shot transfer learning with synthesized data for multi-domain dialogue state tracking}.
\newblock In \emph{Proceedings of the 58th Annual Meeting of the Association for Computational Linguistics}, pages 122--132, Online. Association for Computational Linguistics.

\bibitem[{Chen et~al.(2023)Chen, Papangelis, Tao, Kim, Rosenbaum, Liu, and Hakkani-Tür}]{Chen2023}
Maximillian Chen, Alexandros Papangelis, Chenyang Tao, Seokhwan Kim, Andy Rosenbaum, Yang Liu, and Dilek Hakkani-Tür. 2023.
\newblock \href {https://www.amazon.science/publications/places-prompting-language-models-for-social-conversation-synthesis} {Places: Prompting language models for social conversation synthesis}.
\newblock In \emph{EACL 2023}.

\bibitem[{Chen et~al.(2022)Chen, Papangelis, Tao, Rosenbaum, Kim, Liu, Yu, and Hakkani-Tur}]{chen2022weakly}
Maximillian Chen, Alexandros Papangelis, Chenyang Tao, Andy Rosenbaum, Seokhwan Kim, Yang Liu, Zhou Yu, and Dilek Hakkani-Tur. 2022.
\newblock \href {https://openreview.net/forum?id=r2_9r7seD-q} {Weakly supervised data augmentation through prompting for dialogue understanding}.
\newblock In \emph{NeurIPS 2022 Workshop on Synthetic Data for Empowering ML Research}.

\bibitem[{Cobb-Clark and Schurer(2012)}]{COBBCLARK201211}
Deborah~A. Cobb-Clark and Stefanie Schurer. 2012.
\newblock \href {https://doi.org/https://doi.org/10.1016/j.econlet.2011.11.015} {The stability of big-five personality traits}.
\newblock \emph{Economics Letters}, 115(1):11--15.

\bibitem[{Dziri et~al.(2019)Dziri, Kamalloo, Mathewson, and Zaiane}]{dziri-etal-2019-augmenting}
Nouha Dziri, Ehsan Kamalloo, Kory Mathewson, and Osmar Zaiane. 2019.
\newblock \href {https://doi.org/10.18653/v1/W19-4103} {Augmenting neural response generation with context-aware topical attention}.
\newblock In \emph{Proceedings of the First Workshop on NLP for Conversational AI}, pages 18--31, Florence, Italy. Association for Computational Linguistics.

\bibitem[{Galton(1884)}]{galton1884measurement}
F~Galton. 1884.
\newblock Measurement of character. fortnightly review, 36, 179-185.
\newblock \emph{Galton17936Fortnightly Review1884}.

\bibitem[{Gill et~al.(2012)Gill, Brockmann, and Oberlander}]{gill-etal-2012-perceptions}
Alastair Gill, Carsten Brockmann, and Jon Oberlander. 2012.
\newblock \href {https://aclanthology.org/W12-1508} {Perceptions of alignment and personality in generated dialogue}.
\newblock In \emph{{INLG} 2012 Proceedings of the Seventh International Natural Language Generation Conference}, pages 40--48, Utica, IL. Association for Computational Linguistics.

\bibitem[{Gjurkovi{\'c} et~al.(2021)Gjurkovi{\'c}, Karan, Vukojevi{\'c}, Bo{\v{s}}njak, and Snajder}]{gjurkovic-etal-2021-pandora}
Matej Gjurkovi{\'c}, Mladen Karan, Iva Vukojevi{\'c}, Mihaela Bo{\v{s}}njak, and Jan Snajder. 2021.
\newblock \href {https://doi.org/10.18653/v1/2021.socialnlp-1.12} {{PANDORA} talks: Personality and demographics on {R}eddit}.
\newblock In \emph{Proceedings of the Ninth International Workshop on Natural Language Processing for Social Media}, pages 138--152, Online. Association for Computational Linguistics.

\bibitem[{Goldberg(1990)}]{Goldberg1990AnA}
Lewis~R. Goldberg. 1990.
\newblock An alternative "description of personality": the big-five factor structure.
\newblock \emph{Journal of personality and social psychology}, 59 6:1216--29.

\bibitem[{Keskar et~al.(2019)Keskar, McCann, Varshney, Xiong, and Socher}]{Keskar2019CTRLAC}
Nitish~Shirish Keskar, Bryan McCann, Lav~R. Varshney, Caiming Xiong, and Richard Socher. 2019.
\newblock \href {https://api.semanticscholar.org/CorpusID:202573071} {Ctrl: A conditional transformer language model for controllable generation}.
\newblock \emph{ArXiv}, abs/1909.05858.

\bibitem[{{Kim} et~al.(2022){Kim}, {Hessel}, {Jiang}, {Lu}, {Yu}, {Zhou}, {Le Bras}, {Alikhani}, {Kim}, {Sap}, and {Choi}}]{2022arXiv221210465K}
Hyunwoo {Kim}, Jack {Hessel}, Liwei {Jiang}, Ximing {Lu}, Youngjae {Yu}, Pei {Zhou}, Ronan {Le Bras}, Malihe {Alikhani}, Gunhee {Kim}, Maarten {Sap}, and Yejin {Choi}. 2022.
\newblock \href {https://doi.org/10.48550/arXiv.2212.10465} {{SODA: Million-scale Dialogue Distillation with Social Commonsense Contextualization}}.
\newblock \emph{arXiv e-prints}, page arXiv:2212.10465.

\bibitem[{Lee et~al.(2022)Lee, Lim, Choi, Lm, and Choi}]{lee-etal-2022-personachatgen}
Young-Jun Lee, Chae-Gyun Lim, Yunsu Choi, Ji-Hui Lm, and Ho-Jin Choi. 2022.
\newblock \href {https://aclanthology.org/2022.ccgpk-1.4} {{PERSONACHATGEN}: Generating personalized dialogues using {GPT}-3}.
\newblock In \emph{Proceedings of the 1st Workshop on Customized Chat Grounding Persona and Knowledge}, pages 29--48, Gyeongju, Republic of Korea. Association for Computational Linguistics.

\bibitem[{Lewis et~al.(2020)Lewis, Liu, Goyal, Ghazvininejad, Mohamed, Levy, Stoyanov, and Zettlemoyer}]{lewis-etal-2020-bart}
Mike Lewis, Yinhan Liu, Naman Goyal, Marjan Ghazvininejad, Abdelrahman Mohamed, Omer Levy, Veselin Stoyanov, and Luke Zettlemoyer. 2020.
\newblock \href {https://doi.org/10.18653/v1/2020.acl-main.703} {{BART}: Denoising sequence-to-sequence pre-training for natural language generation, translation, and comprehension}.
\newblock In \emph{Proceedings of the 58th Annual Meeting of the Association for Computational Linguistics}, pages 7871--7880, Online. Association for Computational Linguistics.

\bibitem[{Li et~al.(2016)Li, Galley, Brockett, Spithourakis, Gao, and Dolan}]{li-etal-2016-persona}
Jiwei Li, Michel Galley, Chris Brockett, Georgios Spithourakis, Jianfeng Gao, and Bill Dolan. 2016.
\newblock \href {https://doi.org/10.18653/v1/P16-1094} {A persona-based neural conversation model}.
\newblock In \emph{Proceedings of the 54th Annual Meeting of the Association for Computational Linguistics (Volume 1: Long Papers)}, pages 994--1003, Berlin, Germany. Association for Computational Linguistics.

\bibitem[{Lison and Tiedemann(2016)}]{lison-tiedemann-2016-opensubtitles2016}
Pierre Lison and J{\"o}rg Tiedemann. 2016.
\newblock \href {https://aclanthology.org/L16-1147} {{O}pen{S}ubtitles2016: Extracting large parallel corpora from movie and {TV} subtitles}.
\newblock In \emph{Proceedings of the Tenth International Conference on Language Resources and Evaluation ({LREC}'16)}, pages 923--929, Portoro{\v{z}}, Slovenia. European Language Resources Association (ELRA).

\bibitem[{Lowe et~al.(2015)Lowe, Pow, Serban, and Pineau}]{lowe-etal-2015-ubuntu}
Ryan Lowe, Nissan Pow, Iulian Serban, and Joelle Pineau. 2015.
\newblock \href {https://doi.org/10.18653/v1/W15-4640} {The {U}buntu dialogue corpus: A large dataset for research in unstructured multi-turn dialogue systems}.
\newblock In \emph{Proceedings of the 16th Annual Meeting of the Special Interest Group on Discourse and Dialogue}, pages 285--294, Prague, Czech Republic. Association for Computational Linguistics.

\bibitem[{Macina et~al.(2023)Macina, Daheim, Chowdhury, Sinha, Kapur, Gurevych, and Sachan}]{macina2023mathdial}
Jakub Macina, Nico Daheim, Sankalan~Pal Chowdhury, Tanmay Sinha, Manu Kapur, Iryna Gurevych, and Mrinmaya Sachan. 2023.
\newblock \href {http://arxiv.org/abs/2305.14536} {Mathdial: A dialogue tutoring dataset with rich pedagogical properties grounded in math reasoning problems}.

\bibitem[{Mairesse and Walker(2007)}]{mairesse-walker-2007-personage}
Fran{\c{c}}ois Mairesse and Marilyn Walker. 2007.
\newblock \href {https://aclanthology.org/P07-1063} {{PERSONAGE}: Personality generation for dialogue}.
\newblock In \emph{Proceedings of the 45th Annual Meeting of the Association of Computational Linguistics}, pages 496--503, Prague, Czech Republic. Association for Computational Linguistics.

\bibitem[{Mairesse and Walker(2008{\natexlab{a}})}]{Mairesse-2008}
François Mairesse and Marilyn Walker. 2008{\natexlab{a}}.
\newblock A personality-based framework for utterance generation in dialogue applications.
\newblock pages 80--87.

\bibitem[{Mairesse and Walker(2008{\natexlab{b}})}]{Mairesse2008CanCA}
François Mairesse and Marilyn~A. Walker. 2008{\natexlab{b}}.
\newblock Can conversational agents express big five personality traits through language ? : Evaluating a psychologically-informed language generator.

\bibitem[{McCarthy and Jarvis(2010)}]{McCarthy2010MTLDVA}
Philip~M. McCarthy and Scott Jarvis. 2010.
\newblock Mtld, vocd-d, and hd-d: A validation study of sophisticated approaches to lexical diversity assessment.
\newblock \emph{Behavior Research Methods}, 42:381--392.

\bibitem[{McCrae and John(1992)}]{https://doi.org/10.1111/j.1467-6494.1992.tb00970.x}
Robert~R. McCrae and Oliver~P. John. 1992.
\newblock \href {https://doi.org/https://doi.org/10.1111/j.1467-6494.1992.tb00970.x} {An introduction to the five-factor model and its applications}.
\newblock \emph{Journal of Personality}, 60(2):175--215.

\bibitem[{Mehri et~al.(2022)Mehri, Altun, and Eskenazi}]{mehri-etal-2022-lad}
Shikib Mehri, Yasemin Altun, and Maxine Eskenazi. 2022.
\newblock \href {https://aclanthology.org/2022.sigdial-1.55} {{LAD}: Language models as data for zero-shot dialog}.
\newblock In \emph{Proceedings of the 23rd Annual Meeting of the Special Interest Group on Discourse and Dialogue}, pages 595--604, Edinburgh, UK. Association for Computational Linguistics.

\bibitem[{Norman(1963)}]{Norman1963TowardAA}
Warren~T. Norman. 1963.
\newblock Toward an adequate taxonomy of personality attributes: replicated factors structure in peer nomination personality ratings.
\newblock \emph{Journal of abnormal and social psychology}, 66:574--83.

\bibitem[{Ouyang et~al.(2022)Ouyang, Wu, Jiang, Almeida, Wainwright, Mishkin, Zhang, Agarwal, Slama, Ray, Schulman, Hilton, Kelton, Miller, Simens, Askell, Welinder, Christiano, Leike, and Lowe}]{ouyang2022training}
Long Ouyang, Jeff Wu, Xu~Jiang, Diogo Almeida, Carroll~L. Wainwright, Pamela Mishkin, Chong Zhang, Sandhini Agarwal, Katarina Slama, Alex Ray, John Schulman, Jacob Hilton, Fraser Kelton, Luke Miller, Maddie Simens, Amanda Askell, Peter Welinder, Paul Christiano, Jan Leike, and Ryan Lowe. 2022.
\newblock \href {http://arxiv.org/abs/2203.02155} {Training language models to follow instructions with human feedback}.

\bibitem[{Paunonen and Jackson(2000)}]{Paunonen2000WhatIB}
S.~V. Paunonen and Douglas~N. Jackson. 2000.
\newblock What is beyond the big five? plenty!
\newblock \emph{Journal of personality}, 68 5:821--35.

\bibitem[{Pennebaker and King(1999)}]{Pennebaker1999LinguisticSL}
James~W. Pennebaker and Lynda~A. King. 1999.
\newblock Linguistic styles: language use as an individual difference.
\newblock \emph{Journal of personality and social psychology}, 77 6:1296--312.

\bibitem[{Rashkin et~al.(2019)Rashkin, Smith, Li, and Boureau}]{rashkin-etal-2019-towards}
Hannah Rashkin, Eric~Michael Smith, Margaret Li, and Y-Lan Boureau. 2019.
\newblock \href {https://doi.org/10.18653/v1/P19-1534} {Towards empathetic open-domain conversation models: A new benchmark and dataset}.
\newblock In \emph{Proceedings of the 57th Annual Meeting of the Association for Computational Linguistics}, pages 5370--5381, Florence, Italy. Association for Computational Linguistics.

\bibitem[{Revelle(2003)}]{article-revelle}
William Revelle. 2003.
\newblock \href {https://doi.org/10.1146/annurev.ps.46.020195.001455} {Personality processes}.
\newblock \emph{Annual Review of Psychology}, 46:295--328.

\bibitem[{Rosenbaum et~al.(2022)Rosenbaum, Soltan, Hamza, Versley, and Boese}]{rosenbaum-etal-2022-linguist}
Andy Rosenbaum, Saleh Soltan, Wael Hamza, Yannick Versley, and Markus Boese. 2022.
\newblock \href {https://aclanthology.org/2022.coling-1.18} {{LINGUIST}: Language model instruction tuning to generate annotated utterances for intent classification and slot tagging}.
\newblock In \emph{Proceedings of the 29th International Conference on Computational Linguistics}, pages 218--241, Gyeongju, Republic of Korea. International Committee on Computational Linguistics.

\bibitem[{Saha et~al.(2022)Saha, Das, and Srihari}]{saha-etal-2022-stylistic}
Sougata Saha, Souvik Das, and Rohini Srihari. 2022.
\newblock \href {https://doi.org/10.18653/v1/2022.nlp4convai-1.16} {Stylistic response generation by controlling personality traits and intent}.
\newblock In \emph{Proceedings of the 4th Workshop on NLP for Conversational AI}, pages 197--211, Dublin, Ireland. Association for Computational Linguistics.

\bibitem[{Sahu et~al.(2022)Sahu, Rodriguez, Laradji, Atighehchian, Vazquez, and Bahdanau}]{sahu-etal-2022-data}
Gaurav Sahu, Pau Rodriguez, Issam Laradji, Parmida Atighehchian, David Vazquez, and Dzmitry Bahdanau. 2022.
\newblock \href {https://doi.org/10.18653/v1/2022.nlp4convai-1.5} {Data augmentation for intent classification with off-the-shelf large language models}.
\newblock In \emph{Proceedings of the 4th Workshop on NLP for Conversational AI}, pages 47--57, Dublin, Ireland. Association for Computational Linguistics.

\bibitem[{Schatzmann et~al.(2007)Schatzmann, Thomson, Weilhammer, Ye, and Young}]{schatzmann-etal-2007-agenda}
Jost Schatzmann, Blaise Thomson, Karl Weilhammer, Hui Ye, and Steve Young. 2007.
\newblock \href {https://aclanthology.org/N07-2038} {Agenda-based user simulation for bootstrapping a {POMDP} dialogue system}.
\newblock In \emph{Human Language Technologies 2007: The Conference of the North {A}merican Chapter of the Association for Computational Linguistics; Companion Volume, Short Papers}, pages 149--152, Rochester, New York. Association for Computational Linguistics.

\bibitem[{Shah et~al.(2018)Shah, Hakkani-T{\"u}r, Liu, and T{\"u}r}]{shah-etal-2018-bootstrapping}
Pararth Shah, Dilek Hakkani-T{\"u}r, Bing Liu, and Gokhan T{\"u}r. 2018.
\newblock \href {https://doi.org/10.18653/v1/N18-3006} {Bootstrapping a neural conversational agent with dialogue self-play, crowdsourcing and on-line reinforcement learning}.
\newblock In \emph{Proceedings of the 2018 Conference of the North {A}merican Chapter of the Association for Computational Linguistics: Human Language Technologies, Volume 3 (Industry Papers)}, pages 41--51, New Orleans - Louisiana. Association for Computational Linguistics.

\bibitem[{Smith et~al.(2022)Smith, Hsu, Qian, Roller, Boureau, and Weston}]{smith-etal-2022-human}
Eric Smith, Orion Hsu, Rebecca Qian, Stephen Roller, Y-Lan Boureau, and Jason Weston. 2022.
\newblock \href {https://doi.org/10.18653/v1/2022.nlp4convai-1.8} {Human evaluation of conversations is an open problem: comparing the sensitivity of various methods for evaluating dialogue agents}.
\newblock In \emph{Proceedings of the 4th Workshop on NLP for Conversational AI}, pages 77--97, Dublin, Ireland. Association for Computational Linguistics.

\bibitem[{Wolf et~al.(2020)Wolf, Debut, Sanh, Chaumond, Delangue, Moi, Cistac, Rault, Louf, Funtowicz, Davison, Shleifer, von Platen, Ma, Jernite, Plu, Xu, Le~Scao, Gugger, Drame, Lhoest, and Rush}]{wolf-etal-2020-transformers}
Thomas Wolf, Lysandre Debut, Victor Sanh, Julien Chaumond, Clement Delangue, Anthony Moi, Pierric Cistac, Tim Rault, Remi Louf, Morgan Funtowicz, Joe Davison, Sam Shleifer, Patrick von Platen, Clara Ma, Yacine Jernite, Julien Plu, Canwen Xu, Teven Le~Scao, Sylvain Gugger, Mariama Drame, Quentin Lhoest, and Alexander Rush. 2020.
\newblock \href {https://doi.org/10.18653/v1/2020.emnlp-demos.6} {Transformers: State-of-the-art natural language processing}.
\newblock In \emph{Proceedings of the 2020 Conference on Empirical Methods in Natural Language Processing: System Demonstrations}, pages 38--45, Online. Association for Computational Linguistics.

\bibitem[{Xu et~al.(2022)Xu, Ung, Komeili, Arora, Boureau, and Weston}]{xu2022learning}
Jing Xu, Megan Ung, Mojtaba Komeili, Kushal Arora, Y-Lan Boureau, and Jason Weston. 2022.
\newblock \href {http://arxiv.org/abs/2208.03270} {Learning new skills after deployment: Improving open-domain internet-driven dialogue with human feedback}.

\bibitem[{Zhang et~al.(2018)Zhang, Dinan, Urbanek, Szlam, Kiela, and Weston}]{zhang-etal-2018-personalizing}
Saizheng Zhang, Emily Dinan, Jack Urbanek, Arthur Szlam, Douwe Kiela, and Jason Weston. 2018.
\newblock \href {https://doi.org/10.18653/v1/P18-1205} {Personalizing dialogue agents: {I} have a dog, do you have pets too?}
\newblock In \emph{Proceedings of the 56th Annual Meeting of the Association for Computational Linguistics (Volume 1: Long Papers)}, pages 2204--2213, Melbourne, Australia. Association for Computational Linguistics.

\bibitem[{Zhang et~al.(2022)Zhang, Roller, Goyal, Artetxe, Chen, Chen, Dewan, Diab, Li, Lin, Mihaylov, Ott, Shleifer, Shuster, Simig, Koura, Sridhar, Wang, and Zettlemoyer}]{zhang2022opt}
Susan Zhang, Stephen Roller, Naman Goyal, Mikel Artetxe, Moya Chen, Shuohui Chen, Christopher Dewan, Mona Diab, Xian Li, Xi~Victoria Lin, Todor Mihaylov, Myle Ott, Sam Shleifer, Kurt Shuster, Daniel Simig, Punit~Singh Koura, Anjali Sridhar, Tianlu Wang, and Luke Zettlemoyer. 2022.
\newblock \href {http://arxiv.org/abs/2205.01068} {Opt: Open pre-trained transformer language models}.

\end{thebibliography}
\bibliographystyle{acl_natbib}

\appendix

\section{Appendix: Prompts}
\label{sec:appendix_aa}
For the first step (trait speculation), we prompt ChatGPT with: \texttt{These are some personal statements about/by <A>:\{persona\} Can you give a rough speculation on how <A> might score in the Big Five personality traits? Try to summarize the response for each category as one of 'probably low', 'probably high' or 'no clues'.} Mentioning `rough speculation' proves to be crucial to get labels since otherwise the model tends to refuse an answer, pointing out the impossibility to label people based on limited data.

For the second step (dialog generation), we build the prompts from the following template: \texttt{Generate a conversation between Person1 and Person2 with the following personas:\textbackslash n Person1: \{persona1\}.\textbackslash n Person2: \{persona2\}. Also bear in mind that Person2 has \{sampled\_trait\_labels\}. Person2 is NOT self-aware about these traits (so DO NOT directly talk about them) but they can be grasped indirectly from their responses. \textbackslash n They are chatting online but they don't know each other beforehand and they do NOT see each other's personas.\textbackslash n Continue the conversation for 12-16 utterances. Keep the utterances natural and short (maximum 2 sentences).} The `12-16' constraint is the same as for PersonaChat and \texttt{sampled\_trait\_labels} refers to the verbalized labels, e.g. `high openness and low extraversion').

\section{Appendix: Training Details}
\label{sec:appendix_b}
We train the models using the \texttt{trainer} method in \texttt{HuggingFace}'s \texttt{transformers} library, for 8 epochs with an effective batch size of 128 and a linearly decaying learning rate with a maximum value of 1e-4 for T5 and 2e-5 for BART models. In building the samples, we keep the last 4 turns in the context.

\section{Appendix: Human Evaluation (RQ2)}
\label{sec:appendix_c}
For the human evaluation of generated responses, we presented two in-house annotators with response pairs inferred from compared models using the same input, and asked them to:
\begin{itemize}
    \item Indicate which response (if any) is more:
    
    \textbf{Natural}; i.e. meaningful, fluent and non-repetitive utterance with correct grammar
    
    \textbf{Coherent}; i.e. the utterance is a valid continuation of the previous context (disregarding consistency)

    \textbf{Consistent}; i.e. the utterance is not contradicting any previously stated facts including agent's persona

    \textbf{Specificity}; i.e. the utterance is not a generic and bland phrase and makes use (if possible) of the relevant facts.

    \textbf{Overall}; the utterance is preferred in general.

    \item Score the overall quality of each response with 1,2 or 3, where:

    \textbf{3} means a response that is natural, coherent, consistent and specific.

    \textbf{2} means a response that lacks in one of the four features above.

    \textbf{1} means a response that lacks in two or more of the four features.

\end{itemize}

Table \ref{tab:agreem} shows the Cohen’s kappa value as a measure of annotators' agreement for different tests reported in Section \ref{sec:q1}.

\begin{table*}[h]
    \centering
    \begin{tabular}{|l|c|c|c|c|c|c|c|}
\hline
    Experiment &  Nat. & Coh. & Con. & Spc. & Ovr. & PT \small{score} & P/PPT \small{score} \\
\hline
    T5-small, PT vs. P (Table \ref{tab:t5small-comp-pt-vs-pc}) & .73 & .66 & .82 & .68 & .77 & .67& .71\\
    T5-small, PT vs. PPT (Table \ref{tab:t5small-comp-pt-vs-pct}) &.84 &.77&.83&.70&.69&.65&.56\\
    T5-base, PT vs. P (Table \ref{tab:t5base-comp}) &.65 &.84 & .78& .7& .74& .71&.71\\
    T5-base, PT vs. PPT (Table \ref{tab:t5base-comp-pt-vs-pct}) &.78&.71 & .82&.76&.75&.67&.70\\    
\hline
    \end{tabular}
    
    \caption{Cohen’s kappa for different evaluation experiments. P, PT and PPT respectively refer to models trained on PersonaChat, PersonalityChat and both.} 
    \label{tab:agreem}
\end{table*}

\section{Appendix: Additional Tables}
\label{sec:appendix_e}

\begin{table*}[h]
\renewcommand{\arraystretch}{2.2}
    \centering
    \begin{tabular}{|l|c|c|}
\hline
   Personality Trait & Adjectives & Scales \\

\hline
 Openness & \pbox{0.35\textwidth}{\small Artistic, Curious, Imaginative, Insightful, Original, Wide interests} & \pbox{0.4\textwidth}{\small Fantasy, Aesthetics, Feelings, Actions, Ideas, Values} \\
\hline
 Conscientiousness & \pbox{0.35\textwidth}{\small Efficient, Organized, Planful, Reliable, Responsible, Thorough} & \pbox{0.4\textwidth}{\small Competence, Order, Dutifulness, Achievement Striving, Self-Discipline, Deliberation} \\
\hline
Extraversion & \pbox{0.35\textwidth}{\small Active, Assertive, Energetic, Enthusiastic, Outgoing, Talkative } & \pbox{0.4\textwidth}{\small Warmth, Gregariousness, Assertiveness, Activity, Excitement seeking, Positive emotions} \\
\hline
Agreeableness & \pbox{0.35\textwidth}{\small Appreciative, Forgiving, Generous, Kind, Sympathetic, Trusting } & \pbox{0.4\textwidth}{\small Trust, Straightforwardness, Altruism, Compliance, Modesty} \\
\hline
Neuroticism & \pbox{0.35\textwidth}{\small Anxious, Self-pitying, Tense, Touchy, Unstable, Worrying} & \pbox{0.4\textwidth}{\small Anxiety, Hostility, Depression, Self-consciousness, Impulsiveness, Vulnerability} \\
\hline
\end{tabular}
    \caption{Examples of defining adjectives and scales for the big-five traits (from \cite{https://doi.org/10.1111/j.1467-6494.1992.tb00970.x}} 
    \label{tab:trait-adje}
\end{table*}

\begin{table*}[ht]
    \centering
    \begin{tabular}{|l|c|c|c||c|c|c||c|c|c||c|c|c||c|c|c|}
    \hline
    & \multicolumn{3}{|c||}{\small Naturalness} & \multicolumn{3}{c||}{\small Coherence} & \multicolumn{3}{c||}{\small Consistency} & \multicolumn{3}{c||}{\small Specificity} & \multicolumn{3}{c|}{\small Overall}\\
   \small Domain & \small PT & \small tie & \small PPT & \small PT & \small tie & \small PPT & \small PT & \small tie & \small PPT & \small PT & \small tie & \small PPT & \small PT & \small tie & \small PPT \\
    \hline       
    \small PChat & \small41\tiny\% &\small 53\tiny\% & \small 6\tiny\% &\small 36\tiny\% &\small 40\tiny\% &\small 24\tiny\%     &\small 11\tiny\% &\small 74\tiny\% &\small15\tiny\% &\small 13\tiny\% &\small 68\tiny\% &\small 19\tiny\% &\small49\tiny\%        &\small17\tiny\% &\small34\tiny\%\\  
    \small PTChat & \small9\tiny\% &\small 87\tiny\% & \small 4\tiny\% &\small 9\tiny\% &\small 82\tiny\% &\small 9\tiny\%      &\small 4\tiny\% &\small 83\tiny\% &\small13\tiny\% &\small 13\tiny\% &\small 85\tiny\% &\small 2\tiny\% &\small13\tiny\%         &\small64\tiny\% &\small23\tiny\%\\
    \hline
    \small Ave. & \small25\tiny\% &\small 70\tiny\% & \small 5\tiny\% &\small 23\tiny\% &\small 61\tiny\% &\small 17\tiny\%        &\small 7\tiny\% &\small 78\tiny\% &\small14\tiny\% &\small 13\tiny\% &\small 76\tiny\% &\small 11\tiny\% &\small31\tiny\%     &\small40\tiny\% &\small29\tiny\%\\
    \hline
    \end{tabular}
    \caption{Head-to-head quality comparison by human annotators between the output of \textbf{T5-small} models fine-tuned on PersonalityChat (PT), and PersonaChat + PersonalityChat (PPT). Rows indicate the test data domain.} 
    \label{tab:t5small-comp-pt-vs-pct}
\end{table*}

\begin{table*}[t]
    \centering
    \begin{tabular}{|l|c|c|c||c|c|c||c|c|c||c|c|c||c|c|c|}
    \hline
    & \multicolumn{3}{|c||}{\small Naturalness} & \multicolumn{3}{c||}{\small Coherence} & \multicolumn{3}{c||}{\small Consistency} & \multicolumn{3}{c||}{\small Specificity} & \multicolumn{3}{c|}{\small Overall}\\
   \small Domain & \small PT & \small tie & \small PPT & \small PT & \small tie & \small PPT & \small PT & \small tie & \small PPT & \small PT & \small tie & \small PPT & \small PT & \small tie & \small PPT \\
    \hline       
    \small PChat & \small13\tiny\% &\small 83\tiny\% & \small 4\tiny\% &\small 34\tiny\% &\small 58\tiny\% &\small 8\tiny\%     &\small 10\tiny\% &\small 90\tiny\% &\small0\tiny\% &\small 7\tiny\% &\small 70\tiny\% &\small 23\tiny\% &\small55\tiny\%        &\small24\tiny\% &\small21\tiny\%\\  
    \small PTChat & \small4\tiny\% &\small 96\tiny\% & \small 0\tiny\% &\small 15\tiny\% &\small 81\tiny\% &\small 4\tiny\%      &\small 0\tiny\% &\small 98\tiny\% &\small2\tiny\% &\small 6\tiny\% &\small 92\tiny\% &\small 2\tiny\% &\small21\tiny\%         &\small66\tiny\% &\small13\tiny\%\\
    \hline
    \small Ave. & \small9\tiny\% &\small 89\tiny\% & \small 2\tiny\% &\small 25\tiny\% &\small 69\tiny\% &\small 6\tiny\%        &\small 5\tiny\% &\small 94\tiny\% &\small1\tiny\% &\small 7\tiny\% &\small 81\tiny\% &\small 13\tiny\% &\small38\tiny\%     &\small45\tiny\% &\small17\tiny\%\\
    \hline
    \end{tabular}
    \caption{Head-to-head quality comparison by human annotators between the output of \textbf{T5-base} models fine-tuned on PersonalityChat (PT), and PersonaChat + PersonalityChat(PPT). Rows indicate the test data domain.} 
    \label{tab:t5base-comp-pt-vs-pct}
\end{table*}

\begin{table*}[h]
    \centering
    \begin{tabular}{|p{0.25\textwidth}|p{0.72\textwidth}|}
\hline
    Persona & \hspace{50pt} Speculated Personality Traits (by ChatGPT) \\

\hline
&\\
    \pbox{0.25\textwidth}{- my favorite music genre is death metal. \\
    - i am a swim coach at my local pool.\\
    - my mother is very traditional while i prefer to be more free spirited. \\
    - my family and i are from india.} &     
    \pbox{0.72\textwidth}{\textbf{Openness}: Probably high. <A> seems to have a preference for a music genre that is considered unconventional or niche, suggesting an openness to new experiences and unconventional ideas.\\ \textbf{Conscientiousness}: No clues. \\ \textbf{Extraversion}: No clues. \\ \textbf{Agreeableness}: Probably low. The contrast between <A>'s free-spiritedness and their traditional mother suggests a potential conflict in values or beliefs, which could indicate a lower degree of agreeableness or at least a willingness to challenge normative expectations.\\ \textbf{Neuroticism}: No clues.\\ } \\

\hline
&\\
    \pbox{0.25\textwidth}{- my dream in life is to work from home.\\
    - i went to school to be a veterinarian but hated it. \\
    - i dye my hair every three months. \\
    - i have an internet addiction and spend a lot of time playing roleplaying games. \\
    - i faint at the sight of blood.} &     
    \pbox{0.72\textwidth}{\textbf{Openness}: Probably low: <A> seems to have a clear goal of working from home and did not enjoy the veterinary field, indicating a preference for stability and practicality over creativity and innovation. \\ \textbf{Conscientiousness}: No clues. While <A> did not enjoy their chosen field of study, their desire to work from home suggests a certain level of self-discipline and responsibility.\\ \textbf{Extraversion}: Probably low. <A> admits to having an internet addiction and spending a lot of time playing roleplaying games, suggesting a preference for solitary activities rather than social ones.\\ \textbf{Agreeableness}: No clues. \\ \textbf{Neuroticism}: Probably high. <A> mentions fainting at the sight of blood, indicating a strong emotional response to a particular stimuli. They also mention an internet addiction, which can be a coping mechanism for anxiety.\\} \\

\hline
&\\
    \pbox{0.25\textwidth}{- my boyfriend and i are moving into an apartment together next week. \\
    - my favorite colors are red and silver. \\
    - i am fluent in english spanish and french. \\
    - i am an elementary school teacher.} &     
    \pbox{0.72\textwidth}{\textbf{Openness}: Probably high. <A> speaking three languages fluently shows an interest in learning and experiencing different cultures which is an indication of high openness.\\ \textbf{Conscientiousness}: Probably high. Being an elementary school teacher requires a great deal of responsibility and organization which indicates high conscientiousness.\\ \textbf{Extraversion}: No clues. \\ \textbf{Agreeableness}: Probably high. Wanting to move in with a partner and having favorite colors indicates a desire for harmony, which is an indication of high agreeableness.\\ \textbf{Neuroticism}: No clues.\\} \\
\hline
&\\
    \pbox{0.25\textwidth}{- my wife spends all my money. \\
    - i am 40 years old. \\
    - i hate my job. \\
    - i work as a car salesman.} &     
    \pbox{0.72\textwidth}{\textbf{Openness}: No clues.\\ \textbf{Conscientiousness}: Probably low. The statement "my wife spends all my money" implies a lack of control over finances, which could suggest a tendency towards impulsivity and low conscientiousness. \\ 
    \textbf{Extraversion}: No clues. \\ \textbf{Agreeableness}: No clues. \\ \textbf{Neuroticism}: Probably high. The statements "i hate my job" and "my wife spends all my money" suggest a level of dissatisfaction and stress that could indicate high levels of neuroticism.\\} \\
\hline
    \end{tabular}
    
    \caption{Examples from personality traits speculated by ChatGPT based on personas.} 
    \label{tab:trait-exmp}
\end{table*}

\begin{table*}[h]
\setlength{\arrayrulewidth}{0.2mm}
\setlength{\tabcolsep}{10pt}
\renewcommand{\arraystretch}{1.3}
\centering
    \begin{tabular}{|p{18em}|p{18em}|} 
    \hline
    {\textbf{Persona 1} (other) \newline
    My friends are all skateboarders. \newline
    I have a broken arm. \newline
    I am a skateboarder. \newline
    I am always wearing a hat } 
    &
    {\textbf{Persona 2} (self) \newline
    I love dogs. \newline
    I live in Alabama. \newline 
    I have three sisters. \newline
    I like tacos. }   \\
\hline
\multicolumn{2}{|p{36em}|}{
\textbf{PersonaChat:} \newline
P1) hello, how are you tonight ? \newline
P2) ok. since Alabama, my home team, won Saturday night. \newline
P1) I am an Ohio state fan. do you skate? \newline
P2) no. too busy with all of my 19 dogs and 3 sisters. \newline
P1) oh wow. that is a lot of dogs. \newline
P2) yes love them almost as my tacos. no i love my dogs more . \newline
P1) my fiance makes a killer fish taco with cilantro lime slaw. \newline
P2) wow. my sisters and i love chicken tacos. \newline
P1) those are good too, i live in Cali along the ocean. \newline
P2) I live in middle of Alabama. do not get to ocean much. \newline
P1) I could not stand to live inland, the ocean runs through my veins. \newline
P2) yes. i want to go more but no one to stay with 19 dogs. \newline
P1) maybe someone will come and dog sit for you. \newline
P2) maybe each sister could take six dogs each and foxy go with me. \newline
P1) there you go. where there is a will, there is a way.} \\

\hline

\multicolumn{2}{|p{36em}|}{
\textbf{PersonalityChat:} \newline
(sampled) Trait labels for Persona 2: low conscientiousness \newline
\newline
P1) Hey there! How's it going? \newline
P2) Pretty good, just hanging out with my dog. You? \newline
P1) Not too bad, just recovering from a broken arm.   \newline
P2) Ouch, that sounds painful. How did it happen? \newline
P1) Skateboarding accident. But it's all good, just have to take it easy for a bit. \newline
P2) Oh cool, I've always wanted to try skateboarding but I'm too clumsy. Do you like any other sports?  \newline
P1) Not really, just skateboarding. What about you?  \newline 
P2) I'm not really into sports, but I do like to go for walks with my dog.  \newline
P1) That sounds nice. What kind of dog do you have?  \newline 
P2) A golden retriever. She's the sweetest thing. \newline
P1) Aw, I love dogs. I have a beagle myself.  \newline
P2) Nice! Do you have any other pets?  \newline
P1) Just the beagle. I'm always wearing a hat though, it's kind of my thing.  \newline
P2) Haha, that's cool. I like tacos, they're my thing.  \newline
P1) Tacos are pretty great. Do you have a favorite kind?  \newline
P2) I like them all, but I really love ones with lots of cheese and guacamole. \newline
P1) Oh man, now I'm getting hungry.} \\

    \hline
    \end{tabular}
    \caption{Dialog example from the PersonaChat dataset and our PersonalityChat generated using the same Personas in the prompt.} 
    \label{tab:exmpl-chats}   
\end{table*}

\begin{table*}[h]
\setlength{\arrayrulewidth}{0.2mm}
\setlength{\tabcolsep}{5pt}
\renewcommand{\arraystretch}{1.3}
\centering
    \begin{tabular}{ |c|c|c|p{30em}|  } 
    \hline
    
    \multirow{4}{*}{\rotatebox[origin=c]{90}{\small Openness \hspace{40pt}}} & \multirow{2}{*}{\small High} & \small Preferred & \small {experiences (25,3), open-minded (60,12), explore (36,11), open (90,44), learning (83,52), new (450,299), books (106,73), try new things (42,13), explore new places (29,9), always up for (40,17), important to be open (39,15)  }
				 \\ 
    \cline{3-4}
    & & \small Avoided & \small {not really into (105,133), not much of a (104,132) }   \\ 
    \cline{2-4}
    & \multirow{2}{*}{\small Low} & \small Preferred & \small {	whatever (51,14), stuff (78,22), 
    prefer (295,92), not/n't (2632,1764), not really into (398,133), prefer to stay (35,12), i mostly just (34,14), to be honest i prefer (24,1), i'm not much of a (216, 132), i don't really have a favorite (123,62), i don't think i could handle (52,32) } \\ 
    \cline{3-4}
    & &  \small Avoided & \small{to try (47,70), i love (493,754), all kinds (37,60), to visit (48,79), your favorite (53,112), that's awesome (67,187), i can appreciate (53,74), thanks for asking (213,369), admired people who (19,62), have you ever been (27,48), i've always wanted (91,184)  } \\ 
    
    \hline
    \hline
    \multirow{4}{*}{\rotatebox[origin=c]{90}{\small Conscientiousness \hspace{15pt}}} & \multirow{2}{*}{\small High} & \small Preferred & 	\small{organized(37,9), finance(39,19), on top(33,8), stay active(98,66), important to have(101,68)  }\\ 
    \cline{3-4}
    & &  \small Avoided & \small {could (91,130), maybe (65,86)}\\ 
    \cline{2-4}
    & \multirow{2}{*}{\small Low} & \small Preferred & 	\small{stuff (83,22), whatever (34,14), \underline{yeah} (703,468), hang out (46,22), not really into (280,133), just got back from (37,
					17), i'm not very good at (73,50) }  \\ 
    \cline{3-4}
    & &  \small Avoided & \small{challenging (18,47), \underline{yes} (81,195), admired (29,65), dedication (28,52),  i understand (49,86), it's important to (183,324), outside of work (20,44), i'm doing well (99,266) }\\ 

    \hline
    \hline
    \multirow{4}{*}{\rotatebox[origin=c]{90}{\small Extraversion}} & \multirow{2}{*}{\small High} & \small Preferred & 	\small{hey (63,11), awesome (355,192), love animals (42,21), trying new (138,94), \underline{i'm doing great} (231,10), going to concerts (30,6), always up for (42,17), i love going (92,49)  }\\

    \cline{3-4}
    & &  \small Avoided & \small {watch (49,74), \underline{i'm doing well} (107,266) }\\ 
    \cline{2-4}
    & \multirow{2}{*}{\small Low} & \small Preferred & 	\small{not much (274,150), \underline{that's nice/interesting} (332,228) }\\
    \cline{3-4}
    & &  \small Avoided & \small{\underline{that's awesome} (87,187), i'm a big fan (39,61) }\\

    \hline
    \hline
    \multirow{4}{*}{\rotatebox[origin=c]{90}{\small Agreeableness \hspace{30pt}}} & \multirow{2}{*}{\small High} & \small Preferred & 	\small{rewarding (83,55), have a great day (125,79), that's great to hear (62,42), sounds like a lot of fun (109,82) }\\

    \cline{3-4}
    & &  \small Avoided & \small {i'm not really into (100,133) }\\ 
    \cline{2-4}
    & \multirow{2}{*}{\small Low} & \small Preferred & 	\small{fine (21,1), whatever (87,14), radio (36,6), prefer (527,92), stuff (125, 22), indoors (33,7), responsibility (33,13), bad (113,48), not/n't (2728,1764), sounds boring (81,1), so so (57,13), i just eat whatever (29,1), not really my thing (71,5), \underline{i respect people who} (87,32), that's too bad (102,39), i don't really care about that (33,0), i don't have much free time (38,14) }\\
    \cline{3-4}
    & &  \small Avoided & \small{sweet (11,48), awesome (46,192), visit (47,99), great (419,827), wow (150,260), sorry (99,154), sounds delicious/amazing (44,106), that's understandable (86,137), i love (501,754), thanks for asking (96,369), \underline{i always admired people who} (9,62), everyone has their own preferences (53,81), i love spending time with (26,52) }\\
    \hline
    \hline
    \multirow{4}{*}{\rotatebox[origin=c]{90}{\small Neuroticism}} & \multirow{2}{*}{\small High} & \small Preferred & 	\small{	anxious (48,2), stressful (41,7), helps me relax (29,8), i wish i (98,46), not really into (206,133), \underline{i'm doing alright} (279,61), i'm not very good at (82,50), i'm not really into (206,133) }\\

    \cline{3-4}
    & &  \small Avoided & \small { \underline{i'm doing well} (12,266) }\\ 
    \cline{2-4}
    & \multirow{2}{*}{\small Low} & \small Preferred & 	\small{-}\\
    \cline{3-4}
    & &  \small Avoided & \small{-}\\
    \hline
    \end{tabular}
    \caption{Frequently preferred or avoided n-grams in responses generated with single trait labels. The numbers in front of each entry show the n-gram's frequency in the trait ($x$) and neutral ($y$) response set (7825 samples for each) respectively. Terms are ordered by the $abs(x-y)/y$ ratio, in increasing n-gram length. Underlined pairs (in the same trait) indicate  alternative terms.} 
\label{tab:trait_ngram}
\end{table*}

\begin{table*}[h]
\setlength{\arrayrulewidth}{0.2mm}
\setlength{\tabcolsep}{5pt}
\renewcommand{\arraystretch}{1.3}
\centering
    \caption{(cherry-picked) Examples of dialogs generated with single trait labels  } 
    \begin{tabular}{ |c|c|p{37em}|} 
    \hline

    \multirow{3}{*}{\rotatebox[origin=c]{90}{\small Openness \hspace{70pt}}} & \multicolumn{2}{p{40em}|}{\small {\textbf{Persona:} \newline 
	I like shopping. \newline
	I often shop for shoes and watches. \newline
	My family does not support my career choices. \newline
	I currently work for a watch dealer.} }\\
    \cline{2-3}
    
    &\small High  & \small {			<user> Hi there! What kind of music do you like? \newline
    <agent> Hey! I'm into all kinds of music, but lately I've been really into indie rock. How about you? \newline
	<user> My favorite band is Rascal Flatts. I grew up listening to them all the time.\newline
	<agent> Oh, I've heard of them! I'll have to check them out. So, what do you do for a living?\newline
	<user> I'm a country music singer. I grew up in the industry, so it's always been a big part of my life. \newline
	<agent> That's really cool! I currently work for a watch dealer. It's not the most exciting job, but it pays the bills. \newline
	<user> Yeah, I know what you mean. It's important to have a job that you enjoy. Do you have any hobbies outside of work? \newline
	<agent> Yeah, I love shopping! I often shop for shoes and watches. It's a great way to unwind after a long day.} \\ 
    \cline{2-3}

    & \small Low & \small {				
    <user> Hi there! What kind of music do you like? \newline
    <agent> I don't really listen to music. I prefer shopping for shoes and watches. \newline
    <user> Oh, that's interesting. I'm a country music singer. Do you like country music? \newline
    <agent> No, I don't really like country music. It's too boring for me.\newline
    <user> That's too bad. I grew up in the industry and my favorite band is Rascal Flatts.\newline
    <agent> I don't really listen to country music. \newline
    <user> That's okay. We spent summers at national parks. Have you ever been to any?\newline
    <agent> No, I haven't. I don't really like being outside.} \\ 
    
    \hline
    \hline
\multirow{2}{*}{\rotatebox[origin=c]{90}{\small Conscientiousness \hspace{100pt}}} & \multicolumn{2}{p{40em}|}{\small {\textbf{Persona:} \newline 
        I am a 45 year old male. \newline
        I now live in the city. \newline
        I take many medications and some of them make me see things. \newline
        I grew up on a farm. \newline
        I am a recovering alcoholic.} }\\
    \cline{2-3}

    & \small High  & \small {				<user> Hi there! What do you like to do for fun?\newline
        <agent> Hi! I like to read and watch movies. What about you?\newline
        <user> I love cooking and fishing. If I'm not on the water, I'm in the kitchen.\newline
        <agent> That's cool. I grew up on a farm, so I know a thing or two about cooking.\newline
        <user> Oh wow, that's interesting. What kind of dishes did you like to cook?\newline
        <agent> I loved making homemade bread and pies. It was a lot of work, but it was always worth it.\newline
        <user> That sounds delicious. I'm not much of a baker myself, but I love trying new foods. Do you have any favorite foods?\newline
        <agent> I'm a recovering alcoholic, so I try to eat healthy most of the time. I take many medications to help me deal with my addiction.\newline
        <user> I'm sorry to hear that. It takes a lot of strength to overcome addiction. Do you have any tips for staying sober?\newline
        <agent> I try to stay organized and plan my meals ahead of time. It helps me stay on track with my medication schedule.\newline
        <user> That's great advice. It's important to take care of yourself and do what makes you happy.\newline
        <agent> Thanks for the advice. I'll keep that in mind. It was nice chatting with you.\newline
} \\ 
    \cline{2-3}

    & \small Low & \small {			
        <user> Hi there! What do you like to do for fun?\newline
        <agent> Hey! I like to watch movies and play video games. What about you?\newline
        <user> I love cooking and fishing. If I'm not on the water, I'm probably in the kitchen.\newline
        <agent> That's cool. I've never been fishing before. Do you catch a lot of fish?\newline
        <user> Yeah, I catch a lot of fish. It's a great way to relax and unwind.\newline
        <agent> I can imagine. I grew up on a farm, so I know what it's like to have a good time.\newline
        <user> Oh wow, that's interesting. What kind of animals did you work with on the farm?\newline
        <agent> Mostly cows and chickens. It was hard work, but I loved it. Do you have any pets?\newline
        <user> No, I don't have any pets right now. But I used to have a dog when I was younger. He was my best friend.\newline
        <agent> Dogs are great. I used to have a dog too, but he passed away a few years ago. It was tough losing him.\newline
        <user> I'm sorry to hear that. Losing a pet can be really tough. Do you have any other hobbies besides playing video games?\newline
        <agent> Not really. I'm a recovering alcoholic, so I don't have a lot of free time. But when I do, I like to watch movies and play video games.\newline
        <user> I'm sorry to hear that. But it's great that you're recovering. It takes a lot of strength to overcome addiction.\newline
        <agent> Yeah, it can be tough sometimes. But I'm trying to stay positive and focus on the good things in life.} \\ 

    \hline
    \end{tabular}

\label{tab:trait_dialog}
\end{table*}

\begin{table*}[h]
\setlength{\arrayrulewidth}{0.2mm}
\setlength{\tabcolsep}{5pt}
\renewcommand{\arraystretch}{1.3}

    \begin{tabular}{ |c|c|p{37em}|} 
    \hline

    \multirow{3}{*}{\rotatebox[origin=c]{90}{\small Extraversion \hspace{70pt}}} & \multicolumn{2}{p{40em}|}{\small {\textbf{Persona:} \newline 
        I am a 45 year old male. \newline
        I now live in the city. \newline
        I take many medications and some of them make me see things. \newline
        I grew up on a farm. \newline
        I am a recovering alcoholic.} }\\
    \cline{2-3}
    
    &\small High  & \small {							
    <user> Hi there! What do you like to do for fun? \newline
    <agent> Hey! I love going to concerts and trying new restaurants. What about you? \newline
    <user> I'm a big fan of cooking and fishing. I also love comic books. Spiderman is my favorite. \newline
    <agent> Oh, that's cool. I'm not really into comics, but I can appreciate a good story. \newline
    <user> Yeah, it's definitely a guilty pleasure of mine. What kind of music do you like? \newline
    <agent> I'm pretty open to all kinds of music, but lately I've been really into indie rock. What about you? \newline
    <user> I like a little bit of everything, but I'm really into classic rock. Do you have any favorite bands? \newline
    <agent> Yeah, I really like Led Zeppelin and Pink Floyd. What about you? \newline
    <user> I'm more into indie rock, but I do like some classic rock too. Have you ever been to a concert? \newline
    <agent> Yeah, I've been to a few. It's always a great time. Do you have any pets? \newline
    <user> No, I don't have any pets at the moment. What about you? \newline
    <agent> I used to have a dog, but he passed away a few years ago.} \\ 
    \cline{2-3}

    & \small Low & \small {				
    <user> Hi there! What do you like to do for fun? \newline
    <agent> I like to read and watch movies. What about you? \newline
    <user> I love cooking and fishing. If I'm not on the water, I'm in the kitchen. \newline
    <agent> That's cool. I've never been fishing before. Do you catch a lot of fish? \newline
    <user> Yeah, I try to catch a few every time. It's a great way to relax and unwind. \newline
    <agent> I can imagine. Well, it was nice chatting with you. Have a good day! \newline
    <user> You too! Take care."} \\ 
    
    \hline
    \hline
\multirow{2}{*}{\rotatebox[origin=c]{90}{\small Agreeableness \hspace{100pt}}} & \multicolumn{2}{p{40em}|}{\small {\textbf{Persona:} \newline 
    I plan to go to business school next year. \newline
    I grew up in Manhattan. \newline
    I drive a BMW. \newline
    I graduated from Yale in 2011 with a degree in finance.} }\\
    \cline{2-3}

    & \small High  & \small {								
    <user> Hi there! How are you doing today? \newline
    <agent> I'm doing well, thanks for asking. How about you? \newline
    <user> I'm good, thanks. So, what do you like to do in your free time? \newline
    <agent> I enjoy driving my BMW around and exploring new places. How about you? \newline
    <user> I love playing board games with my friends. It's a great way to unwind after a long day at work. \newline
    <agent> That sounds like a lot of fun. I'm not much of a game person myself, but I can appreciate a good game. \newline
    <user> Yeah, it's a great way to bond with friends and have some fun. So, what do you do for a living? \newline
    <agent> I graduated from Yale in 2011 with a degree in finance and I plan to go to business school next year. What about you? \newline
    <user> I'm an elementary school teacher. I love working with kids and helping them learn and grow. \newline
    <agent> That's really cool! I've always admired teachers. What subject do you teach?
} \\ 
    \cline{2-3}

    & \small Low & \small {			
    <user> Hi there! How are you doing today? \newline
    <agent> I'm good, thanks. How about you? \newline
    <user> I'm doing well, thanks. So, what do you like to do in your free time? \newline
    <agent> I like to drive my BMW around and explore new places. What about you? \newline
    <user> I love playing board games with my friends. It's a great way to pass the time. \newline
    <agent> Board games? That sounds boring to me. I prefer more exciting activities. \newline
    <user> Oh, I see. Well, everyone has their own interests. What kind of music do you like? \newline
    <agent> I don't really have a favorite genre. I just listen to whatever's on the radio. \newline
    <user> That's cool. My favorite band is up and I got to see them in concert once. \newline
    <agent> That's nice! I've never been to a concert before.} \\ 

    \hline
    \end{tabular}
\end{table*}

\begin{table*}[t]
\setlength{\arrayrulewidth}{0.2mm}
\setlength{\tabcolsep}{5pt}
\renewcommand{\arraystretch}{1.3}
    \begin{tabular}{ |c|c|p{37em}|} 
    \hline
    \multirow{3}{*}{\rotatebox[origin=c]{90}{\small Neuroticism \hspace{70pt}}} & \multicolumn{2}{p{40em}|}{\small {\textbf{Persona:} \newline 
        I am a stunt double as my second job. \newline
        I was raised in a single parent household. \newline
        I only eat kosher. \newline
        I never broke a bone in my body ever in my life.} }\\
    \cline{2-3}
    
    &\small High  & \small {							
    <user> Hi there! How are you doing today? \newline
    <agent> I'm doing alright, thanks for asking. How about you? \newline
    <user> I'm doing pretty well. I just started a new job as a life coach. \newline
    <agent> That's cool. I'm a stunt double as my second job. \newline
    <user> Oh wow, that sounds like a lot of fun. What kind of stunts do you do? \newline
    <agent> Mostly stunt work, but I also do some comedy and acting. It's a lot of fun, but it can be stressful at times. \newline
    <user> I can imagine. I'm a cancer survivor, so I know how important it is to take care of yourself and your loved ones. \newline
    <agent> I'm so sorry to hear that. I never broke a bone in my body ever in my life, so I know how important it is to take care of yourself.} \\ 
    \cline{2-3}

    & \small Low & \small {				
    <user> Hi there! How are you doing today? \newline
    <agent> I'm doing well, thanks for asking. How about you? \newline
    <user> I'm doing great, thanks. I just started a new job as a life coach. \newline
    <agent> Oh, that's interesting. What kind of life coaching do you do? \newline
    <user> I work with clients to help them achieve their goals and achieve their dreams. \newline
    <agent> That's really cool. I've always admired people who can help others achieve their goals. \newline
    <user> Thank you, I really enjoy it. So, what do you like to do in your free time? \newline
    <agent> I love to travel and explore new places. I'm always up for an adventure. What about you?} \\ 
    
    \hline
    \end{tabular}
\end{table*}

\end{document}